\documentclass[runningheads]{llncs}
\usepackage{graphicx}
\usepackage{amsmath}
\usepackage{multirow} 
\usepackage{booktabs}
\usepackage{array}
\usepackage{comment}

\usepackage{amsmath}
\usepackage{multirow, makecell, colortbl, hhline}
\usepackage{xcolor}
\usepackage{cellspace}
\usepackage{hyperref}

\begin{document}
%
\title{EAMDrift: An interpretable self retrain 
model for time series\thanks{Supported by organization 1, 2.}}
%
%
\author{Gonçalo Mateus\inst{1, 2}\orcidID{0000-0003-4744-8016} \and
Cláudia Soares\inst{1}\orcidID{0000-0003-3071-6627} \and
João Leitão\inst{1}\orcidID{0000-0001-7916-980X} \and
António Rodrigues\inst{2}}
%
\authorrunning{G. Mateus et al.}
%
\institute{NOVA School of Science and Technology, Lisbon, Portugal \\
\url{https://www.fct.unl.pt/en} 
\and
Novobanco, Lisbon, Portugal\\
\url{https://www.novobanco.pt/}
}
\maketitle              
\begin{abstract} 

The use of machine learning for time series prediction has become increasingly popular across various industries thanks to the availability of time series data and advancements in machine learning algorithms. However, traditional methods for time series forecasting rely on pre-optimized models that are ill-equipped to handle unpredictable patterns in data.

In this paper, we present EAMDrift, a novel method that combines forecasts from multiple individual predictors by weighting each prediction according to a performance metric. EAMDrift is designed to automatically adapt to out-of-distribution patterns in data and identify the most appropriate models to use at each moment through interpretable mechanisms, which include an automatic retraining process. Specifically, we encode different concepts with different models, each functioning as an observer of specific behaviors. The activation of the overall model then identifies which subset of the concept observers is identifying concepts in the data. This activation is interpretable and based on learned rules, allowing to study of input variables relations. 

Our study on real-world datasets shows that EAMDrift outperforms individual baseline models by 20\% and achieves comparable accuracy results to non-interpretable ensemble models. These findings demonstrate the efficacy of EAMDrift for time-series prediction and highlight the importance of interpretability in machine learning models. 


\end{abstract}
\section{Introduction}

Nowadays, vast amounts of time series data are generated and collected from various sources in a streaming setting. Novel algorithms and hardware enable extracting valuable insights from these data streams through machine learning algorithms in fields such as finance~\cite{Cheng22} and public health~\cite{Petropoulos20}. Furthermore, the SARS-CoV-2 pandemic has highlighted the importance of time series prediction in forecasting, such as the spread of infectious diseases and the demand for medical supplies and services~\cite{hyman2021data,bertozzi2020challenges}.

However, the widespread implementation of artificial intelligence is hindered by a lack of trust in multiple industries due to the absence of clarity on the model behavior to back up decisions~\cite{jan2020ai,lopardo2021smace,stiglic2020interpretability,saadallah2022explainable}. Confidence and interpretability are closely connected, and as a public concern, the European ethics guidelines for trustworthy AI~\cite{ai2019high} state that "the degree of interpretability needed is highly dependent on the context," but recommend having "transparent AI systems that explain their decisions to those directly and indirectly affected."

%

As a general concept, time series data is a sequence of unpredictable and varying patterns that evolve. These patterns, which we call "concepts" in this work, are often characterized by high seasonalities~\cite{Petropoulos20,lovell1963seasonal,jiang2013cloud}. Existing approaches typically rely on a single model trained on pre-defined assumptions based on past data~\cite{ariyo2014stock,calheiros2014workload,gardner1985exponential,nsabimana2022forecasting,jayakumar2020self,wen2022transformers,chang2023tdstf,cantero2023convolutional,li2019ea}. While such approaches can produce good results in some instances, different models may yield better estimations for different concepts, as demonstrated by various ensemble modeling approaches \cite{larrea2021extreme,iqbal2019adaptive,jun2023time,8457781,Saadallah19,rossi2014metastream,krawczyk2017ensemble}.
Furthermore, many existing approaches do not consider external factors. For example, relying solely on established concepts to predict future stock trends can be risky in the stock market. Stocks are influenced by politics, events, and investor sentiments \cite{mahmood2014impact,siregar2019impact}.

\begin{figure}
\centering
\includegraphics[width=1\textwidth]{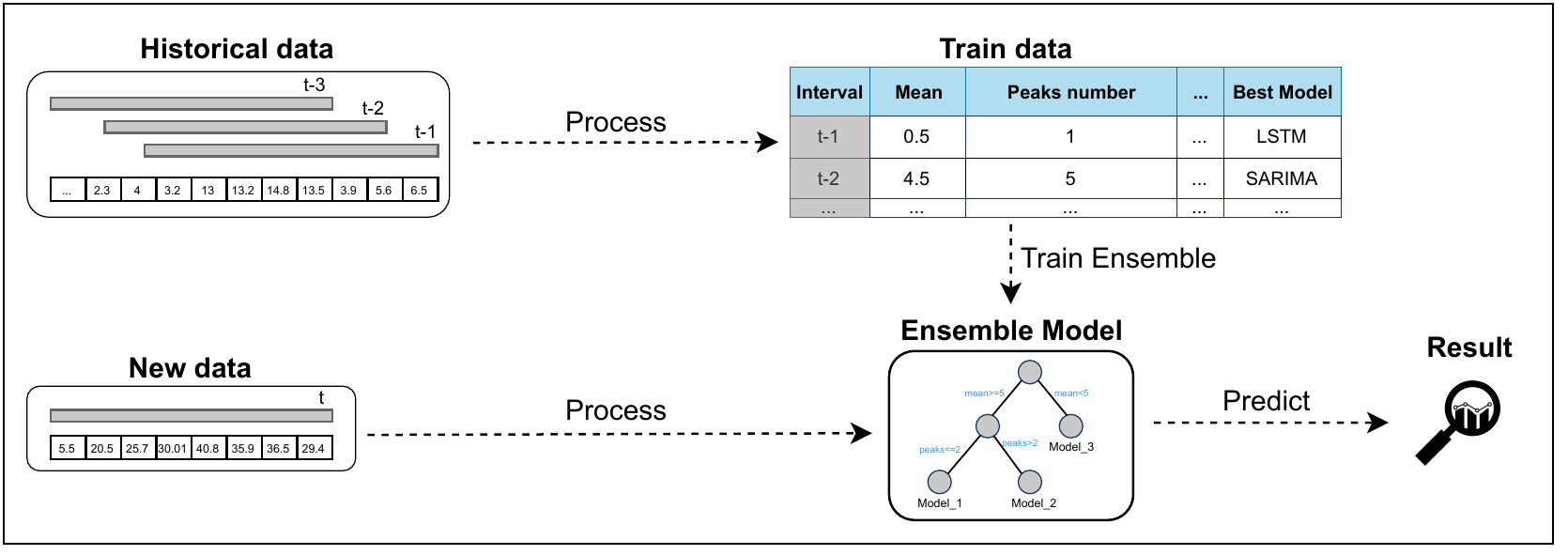}
\caption{Proposed model overview. For each historical window, the model extracts statistics and finds the best model to create a structured table to train the ensemble model.} \label{fig:split_geral_schema_simple}
\end{figure}

Motivated by the need for interpretability, handling of different concepts, and external factors, this paper proposes a novel machine-learning method to forecast time series called the Ensemble Adaptive Model with a Drift detector (EAMDrift). EAMDrift combines the power of multiple individual predictors, such as Prophet, ARIMA, and LSTM, through an interpretable model.

The key idea is to use different models to encode different concepts, each observing specific behaviors. The activation of EAMDrift recognizes concepts in data and assigns weights to each observer for each prediction. These weights are calculated at run time and combined with the observer's predictions to assemble the final result. Additionally, EAMDrift accepts external covariates, allows the study of relations between input variables, and contains a self-retrain mechanism that helps the model adapt to unexpected concepts over time.

As shown in Figure~\ref{fig:split_geral_schema_simple}, EAMDrift generates various splits from historical data and, for each split, extracts a handful of statistics and tests different models to create a structured table. This table serves as the training data for the ensemble model, with extracted statistics and external covariates as the input ($X$) and the best model found as the output ($Y$). Based on interpretable learned rules, this ensemble model assigns weights to each predictor, determining their contribution to the final prediction.

Experimental evaluations using different real-world datasets demonstrate that our model outperforms single approaches and achieves on-par results compared to non-interpretable ensemble models.

The main contributions of this paper are:
\begin{itemize}
    \item A interpretable ensemble method that selects the bests predictors at each point, identifying relevant concepts. 
    \item A method based on statistics to be easier to interpret;
    \item A model that accepts past covariates\footnote{Our model can automatically add covariates related to dates.} (either numerical or categorical) and allows studying relations between them. 
    \item A retrain method that strategically finds potential points to retrain.
\end{itemize}

The rest of this paper is organized as follows: In Section~\ref{sec_related_work}, we presented the Related Work. Subsequently, in Section~\ref{model_architecture}, we present the detailed EAMDrift model architecture. The experimental methodology and datasets used to test our model are presented in Section~\ref{sec_experimental_methodology}, and the respective results are presented in Section~\ref{sec_experimental_results}. Finally, in Section~\ref{sec_conclusion}, we discuss the conclusion and future work.
\section{Related Work}
\label{sec_related_work}

The \textit{``one-fits-all"} style, where a single predictive model is used, has been the most popular technique due to its simplicity and good prediction power~\cite{bankole2013cloud}. These models rely on regression, machine learning, and time series techniques. However, although time series models such as ARIMA~\cite{ariyo2014stock,calheiros2014workload} and variations of Exponential Smoothing~\cite{gardner1985exponential,nsabimana2022forecasting} are the most commonly used methods due to their ability to detect seasonality and cyclic behaviors and their ease of use, they fail when dealing with unpredictable concepts in data. As a result, some works employ machine learning models such as LSTM, CNN, and Transformers. Although training these models requires more effort, they can learn long-term and complex concepts \cite{jayakumar2020self,wen2022transformers,chang2023tdstf,cantero2023convolutional,li2019ea}.

However, research has shown that even with more complex models, a single model cannot handle all unpredictable concepts in data. Therefore, different ensemble and adaptive models have been proposed~\cite{boulegane2020streaming,britto2014dynamic,boulegane2019arbitrated,bifet2009adaptive,saadallah2021actor}.

For different concepts, different models yield better estimations~\cite{liu2017adaptive,MOABook2018}. To address this issue, Iqbal \emph{et al.} proposed a novel adaptive method that automatically identifies the most appropriate model for specific scenarios~\cite{iqbal2019adaptive}. They used classical machine learning methods like LR, SVM, and GBT to forecast data.

Jungang \textit{et al.} proposed a combined Prophet-LSTM method to leverage time series features, such as trend and periodicity, while learning long-term concepts in data~\cite{jun2023time}. The algorithm obtains the final result through linear weighting of the two models.

Kim \textit{et al.} proposed a method called CloudInsight, inspired by the mixture of experts problems~\cite{jacobs1991adaptive}, which assigns weights to each predictor to forecast data~\cite{8457781}.

There have also been efforts to employ interpretability in time series predictions~\cite{saadallah2021explainable}. Some of the frequently used models include linear regression, logistic regression, and the decision tree due to their internal transparency~\cite{molnar2020interpretable}. In other way, drifts can also add interpretability by sounding an alarm when changes in the data are detected~\cite{mayaki2022autoregressive,kolter2005using}.

%
%
\section{EAMDrift: Proposed model architecture}
\label{model_architecture}

Using an innovative model architecture, EAMDrift combines forecasts from multiple individual predictors by weighting each prediction according to a performance metric. EAMDrift identifies the most promising predictors for each concept and assigns them higher weights. By defining a correspondence between models and concepts, we can view each predictor in the ensemble as an observer of the data, indicating the presence of a given pattern by the strength of its prediction.

\begin{figure}
\centering
\includegraphics[width=1\textwidth]{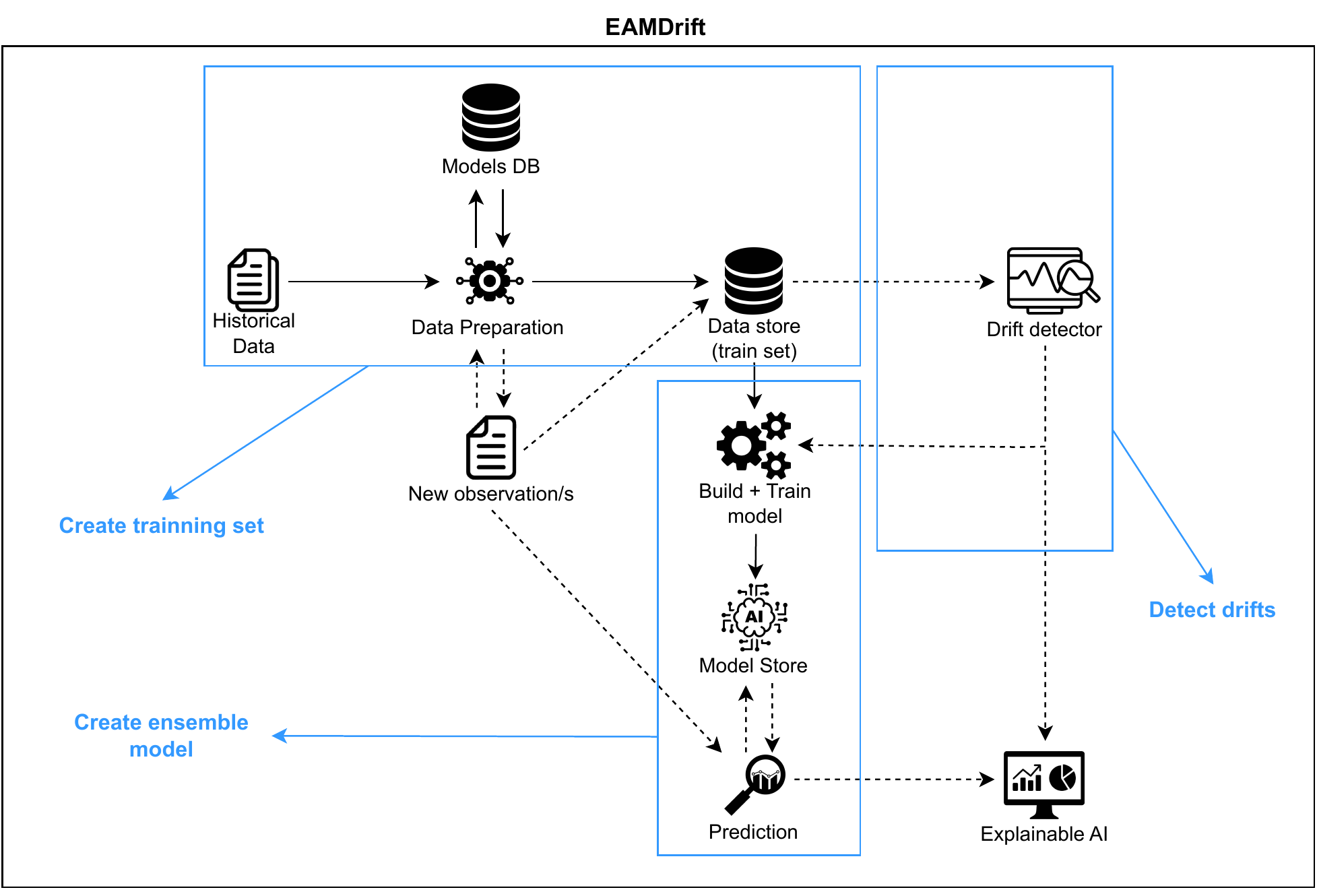}
\caption{Proposed model architecture. The three main components of our proposed model are highlighted in blue.} \label{fig:model_architecture}
\end{figure}

The architecture of our proposed model is depicted in Figure \ref{fig:model_architecture} and will be described next.

\begin{itemize}
    \item In the first step (\textbf{Create training set}), the model starts by using historical workload and covariates data to create a new training set that will be used to train our ensemble model. This historical data will be split into different sliding windows, each with the same size. Then, for each sliding window, we will: run different models (previously selected by the user) and find the best, extract statistics about that window (like mean, number of peaks, and others), and pre-process covariates data (i.e., for categorical covariates the model chooses the most frequent and for numerical the model sums that values for each window). Ultimately, we will have a train set with the columns: statistics, processed covariates, and best model. Each row of the train set will represent a sliding window. Sliding windows encode concepts, and the best model for each concept represents an expert on a given concept.
    \item In the second step (\textbf{Create ensemble model}), the model will use the training set previously created and build an ensemble model. This ensemble model uses RuleFit, an interpretable machine learning model based on rules, and will have as input the columns referring to statistics and processed covariates and as output the column with the best model. In each prediction, the RuleFit model will output probabilities for the selection of each model, and the final prediction will be a combination of the output of different models multiplied by their respective probabilities. Unlike the usual time-series models, our model does not use the time-series points directly to select the weights for each model but instead uses statistics-related data.
    \item The third step (\textbf{Detect drifts}) occurs during real-time usage. When new points are given to the model, the model processes these data to create an input in the necessary format for the ensemble model. At the same time this processing is done, the model tests if these new points trigger the drift detector to query the need to retrain the model. In the positive case, the model retrains before giving the respective predictions.
\end{itemize}

\subsection{Create initial training set}

Before preparing the training set, the user must create the Models Database. This database contains the models the user has chosen to use during training. This model works with any model as long as it is implemented in the code. The model will receive historical data to be trained as input with this definition.
The model expects this data to be a time series DataFrame with a column for dates, another for the variable we want to predict, and the rest regarding covariates (either numerical or categorical) that the user wants to add. Next, the pre-processing step will begin, and the historical input will be divided into different splits and fixed ranges of points.

The user must define the number of $n$ training points and $m$ prediction points. The user can also select the number of splits they want to create to train the model. Suppose the user does not choose any number of splits. In that case, the model will automatically find the maximum number of splits that can be created depending on the values of $n$ and $m$ variables.

\begin{figure}
\centering
\includegraphics[width=1\textwidth]{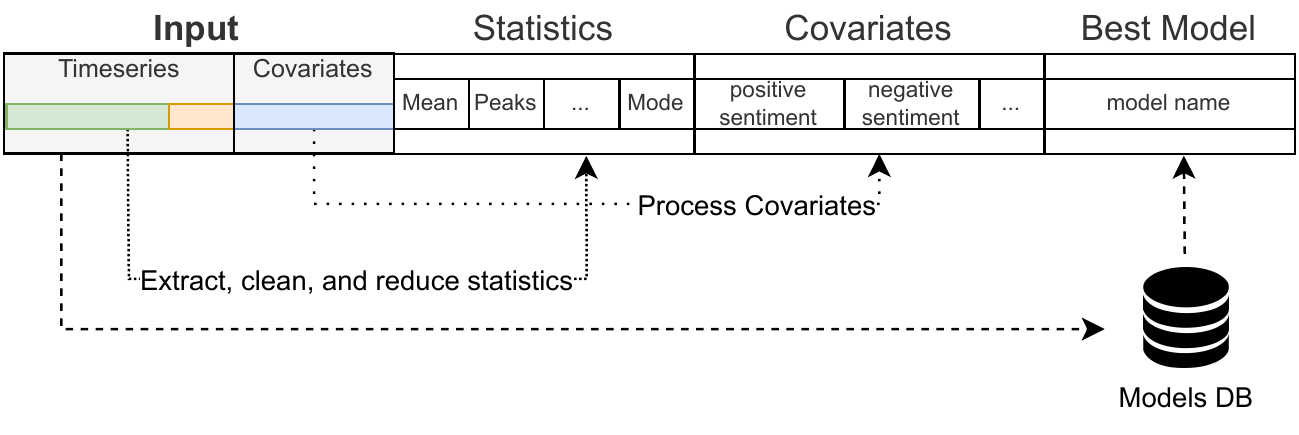}
\caption{Schema of each training set line.} \label{fig:split_geral_schema2}
\end{figure}

With the split ranges defined, the next step will be to process each split, as depicted in Figure~\ref{fig:split_geral_schema2}. For each split, we will extract statistics from the training points, test models from the Models DB using the training and prediction points to validate forecasts, and finally process the covariates. 

The statistics extraction was performed using the Python library \texttt{tsfresh} \cite{christ2018time}, which can automatically compute numerous features of time series data. \texttt{tsfresh} extracts around 800 statistics, ranging from simple and well-known ones to more complex ones. However, handling such a large number of statistics would make the model less interpretable, as it would be too many statistics to process. Therefore, we used data pre-processing and feature reduction techniques to reduce the number of features.

The model first removes columns with more than 50\% null values, columns with more than 95\% feature similarity in values and variance, and correlated columns. On average, these steps reduced the number of statistics from 800 to 250. However, this still remained a very high number to analyze, so we performed an additional feature selection technique based on ElasticNet regression.

ElasticNet is a linear regression model that combines the properties of Lasso and Ridge Regression by adding L1 and L2 regularization terms. ElasticNet minimizes the objective function depicted in equation \eqref{eq:ridge_regression}, where the first term is the MSE loss, and the second term is the L1 regularization term which encourages sparsity (more features being set to zero) by adding a penalty for non-zero weights. The third term is the L2 regularization term which encourages small weights by adding a penalty for large weights. Here, $y$ represents the target sample, $w$ represents the weight vector, $n$ is the number of samples, and $p$ is the number of features.

ElasticNet has two parameters to be defined, the $l1\_ratio$ ($\rho$ in the equation) and the $\alpha$, to balance L1 and L2 regularization. A larger value for alpha means stronger regularization, and a larger value for $l1\_ratio$ means more L1 regularization, which leads to sparsity (if $l1\_ratio=1$, only L1 regularization is used, and if $l1\_ratio=0$, only L2 regularization is used). Our objective in using ElasticNet is to select the most important features while simultaneously shrinking the less important, leading to a more interpretable model. We selected $l1\_ratio$ and $\alpha$ to be 0.7 and 0.9, respectively, leading to an average feature reduction from 250 to 45.

\begin{equation}\label{eq:ridge_regression}
\begin{aligned}
L_{enet}(w) &= \frac{1}{2n} ||y - X_w||^2_2 + \alpha \rho ||w||_1 + \frac{\alpha(1-\rho)}{2} ||w||^2_2
&&\\
&=  \frac{1}{2n} ||y - X_w||^2_2 + \alpha \rho \sum_{j=1}^{p} |w_j| + \frac{\alpha(1-\rho)}{2} \sum_{j=1}^{p} w_j^2
\end{aligned}
\end{equation}

To test models, we use the training points for each split to train every model present in the Models DB and predict the following $m$ points. Then, we register the model that provides the best accuracy, as demonstrated in Figure~\ref{fig:split_geral_schema2}.

The format covariates step is the simplest of the three, and its objective is to aggregate the $n$ covariate points into one line. For numerical covariates, we sum the values of all points, whereas for categorical covariates, we count the most common category in that period. After encoding with the $OneHotEncoder$ function from \texttt{scikit-learn}, the most common category will also be numerical. Our model can also automatically add time covariates such as day, month, and year. If the user wants to use these covariates for each sliding window, our model adds the date from the last training point.

Ultimately, we will have a training set with multiple rows, where each row corresponds to a sliding window. For each line, we will have columns regarding statistics, covariates, and the best model for the respective window, as shown in Figure~\ref{fig:split_geral_schema2}.

\subsection{Creating the ensemble model}

With the training set created, the model is ready to train the ensemble part. We will train a model whose main objective is to provide probabilities for the models present in the Models DB. This model will receive columns regarding statistics and covariates from the training set and will output probabilities for each model based on the best model column. We have selected the RuleFit framework~\cite{friedman2008predictive} and adapted the output to solve our probability task (original RuleFit solves regression tasks). RuleFit was chosen because it is an interpretable model that generates a set of human-readable rules that can be used to make predictions.

After training, the overall model will be stored to predict new points in the future. When a new point appears that needs to be predicted, this point will be added to the data and processed to create a new line similar to the ones created for the training set, but without the \textit{best model} column. This new line will have the same format as the ones given to our RuleFit model for training so that our model can output a prediction. The output of our RuleFit will be a vector $Y$ in the format $n$ (\ref{eq:output_of_ensemble}). $n$ corresponds to the size of the Models DB, and each line of the matrix will contain the probability of use for each model. The probability values will be between 0 and 1, where a value closer to 1 indicates that it will have more impact on the final prediction.
\begin{equation}\label{eq:output_of_ensemble}
\begin{aligned}
Y = \left( \begin{array}{c}
p_1 \\
p_2 \\
... \\
p_n 
\end{array} \right)
\end{aligned}
\end{equation}
With the matrix $Y$ obtained, our overall model will run every model present in Model DB with the training points of the split that is being predicted, creating a new matrix $P$ with the same format of $Y$, but now with the model's predictions. The final prediction will be computed based on equation \eqref{eq:final_prediction}, where the matrices $P$ and $Y$ will be multiplied. 

\begin{equation}\label{eq:final_prediction}
\begin{aligned}
pred(Y, P) &= \sum_{i=1}^{n} Y_i \cdot P_i
\end{aligned}
\end{equation}

\subsection{Drift detector}

The final part of our model is the Drift detector. This component will try to find changes in the data distribution that may indicate the need for retraining. When a new point arrives in the model, the drift detector will run, and if needed, the model will be retrained before giving the prediction \cite{gama2014survey,saadallah2020drift,soares2015dynamic}. Moreover, this will also warn the user to be more careful when analyzing the workload in the following days.

For this paper, we focused on concept drifts, a type of drift occurring when the relationships between the input and output variables change. Concerning our work, a concept drift detector would trigger when the workload trend suffers a change. We tested two concept drift detectors, ADaptive WINdowing (ADWIN) and Kolmogorov-Smirnov Windowing (KSWIN). The drift that had an overall better performance on tests and was thus selected was KSWIN, as it was the one that detects drifts in more critical zones (zones with abrupt changes). KSWIN is based on the Kolmogorov–Smirnov (KS) statistical test that compares a part of data to a reference distribution. It is based on the maximum difference between some sample empirical Cumulative Distribution Function (CDF) and the reference CDF. 
The CDF encodes the probability of a random variable $X$, with a given probability distribution, being found at a value less than or equal to $u$ (Equation \eqref{eq:cdffunct})~\cite{deisenroth2020mathematics}. 

\begin{equation}\label{eq:cdffunct}
\begin{aligned}
CDF_x(u) = P(X \leq u) 
\end{aligned}
\end{equation}

Our method defines an interval to avoid successive retrains in zones where the workload changes significantly. If this interval is defined as 14 days, the model must wait at least 14 days to retrain. Thus, if drifts are detected on 1, 7, and 20 days of a month, the model will retrain in the 1 and 20 days.

\subsection{Model Interpretability}

As mentioned earlier, the interpretability in EAMDrift does not explain the prediction itself. Instead, it is in the ensemble and allows us to point to a concept mediated by the model's choice. We can observe which rules and features contribute more to the prediction. EAMDrift lets us know which models contribute to a specific prediction and what weight was attributed to each. This model uses a combination of different predictors for the final forecast. Furthermore, each rule has a \texttt{support} parameter related to the number of points that satisfy a rule. Rules with more significant values will also help us understand the relations between input variables in our data.

EAMDrift also has an implemented drift detector. Although it is meant to detect moments of potential retraining, it also warns the user to be more attentive to data in the following periods.

\subsection{Details and implementation of EAMDrift}

EAMDrift is compatible with any model as long as the user implements it. If the user selects a model X and during training, it is chosen as the best for less than or equal to one prediction, it will be discarded from the pool of predictors because it did not classify enough points. However, if drift is detected later and the model needs to be retrained, this model X will have the opportunity to enter the predictor pool again.

Our model also allows for selecting a restricted number of predictors in each prediction. For example, if we have six models in the Models DB and select three predictors in each prediction, the model will only use the three predictors with the highest probability in the matrix P (see~\eqref{eq:output_of_ensemble}). To predict subsequent periods, the user can also tune other parameters, such as the number of training points and covariates. Additionally, the user can choose between Mean Square Error (MSE), Mean Absolute Error (MAE), or Mean Absolute Percentage Error (MAPE) as the metric for selecting the best models during the creation of the training set. Finally, the user can define a length for the training set, which restricts the number of points used in each retrain to avoid excessive training set rows.
EAMDrift was implemented in \texttt{Python} (version 3.9.12) on \texttt{Anaconda} using mainly \texttt{Pandas}, \texttt{NumPy}, and \texttt{scikit-learn} libraries. For the machine learning models, we used mainly \texttt{River}, \texttt{Darts}, and \texttt{statsmodels} libraries. The training and inference of EAMDrift were executed in a 6-core Intel(R) Core(TM) i5-8500 CPU @ 3.00GHz. The creation of the training set was parallelized through the function $Parallel$ from \texttt{joblib} library to reduce the processing time. We used six jobs to take maximum advantage of the computer's cores.

\section{Experimental Methodology}
\label{sec_experimental_methodology}

This section outlines the methodology and experimental procedures for validating our proposed model. To this end, we describe the datasets selected for this study and the methodology chosen for conducting the experiments, which includes the setup of EAMDrift parameters.

\subsection{Datasets}

To test EAMDrift, we used five different datasets from different backgrounds. As one of the innovations of our model is the possibility to use and study relations between input variables, we added covariates to each dataset. We sliced a subset of each dataset to avoid creating predictions with excessive points, ensuring that each slice contained at most 5000 points. Moreover, all the datasets were previously processed, and the variable to predict was standardized. 

The first two datasets, \textbf{NB1} and \textbf{NB2}, are private and correspond to the CPU usage of real servers that manage a company's main application. The time step in data is in days and comprises 1460 records. As this corresponds to a real application, we extracted tweets related to the company and ran a sentiment analysis to be used as covariates. The third dataset is the Google Cluster Trace (\textbf{GCT}) \cite{wilkes2011more}. It corresponds to a set of workload data from different servers for May 2019 (for this work, just one server was used). The data has a 5-minute step for the entire month of May and provides CPU and Memory metrics. We grouped data in hours, giving us 744 entries, and memory was used as a covariate. The fourth dataset is the Electric Power Consumption (\textbf{EPC}) \cite{ihepcdata}. It measures the electric power usage in different houses in the zone of Paris, France, and for our test, just one house was chosen. The data has a 1-minute step for nearly four years but was aggregated in hours, giving us 35063 entries. As electric consumption can be related to weather, we used data from ``\texttt{AWOS}" sensors available in \cite{asosmetardata} to be used as covariates. The last dataset used is the Microsoft Stock Value (\textbf{MSV}) \cite{microsoftstockdata}, which consists of Microsoft stock close value from 2006 to 2016 with a 1-day time step. This dataset also includes a sentiment analysis of the financial news related to Microsoft for that period, which will be used as covariates.


\subsection{Experiments Setup}
\label{experimenta_setup_details}

Our model was designed to work with any model, so to test, we selected as predictors five models with different backgrounds, including Prophet, Exponential smoothing (ES), Seasonal Autoregressive Integrated Moving Average (SARIMA), Long-Short Term Memory (LSTM), and Transformer. 
The rest of the EAMDrift parameters were defaulted and can be consulted in the code repository.

To compare validation results, we evaluated our proposed model against single-model approaches. In addition, we created two non-interpretable variations of our model (based on some state-of-the-art works (Section \ref{sec_related_work})), in which we replaced the interpretable ensemble part (RuleFit) of EAMDrift with the Random Decision Forest (RDF) and Support Vector Machine (SVM) models. The purpose of these variations was to compare the effectiveness of RuleFit with other black-box models acting as our ensemble selector.

For each dataset, we used 40\% of the points to train and the remaining to forecast. We elected to utilize only 40\% of the available data points for training, motivated by two primary factors. Firstly, some of the datasets contained a substantial number of observations, and secondly, to test the efficacy of our retrain mechanism. To obtain the results, we used blocked cross-validation, where a default set range is defined for training. That range is used in all iterations of the cross-validation folds. Each prediction was made by training the model with data until the prediction day. The Mean Absolute Percentage Error (MAPE) \eqref{eq:mape} was calculated for each fold and then averaged to produce the final score for each model to assess predictions. 

\begin{equation}\label{eq:mape}
\begin{aligned}
MAPE = \frac{1}{n} \sum_{i=1}^{n} \left| \frac{Actual_i - Predicted_i}{Actual_i} \right| \times 100%
\end{aligned}
\end{equation}

\section{Experimental Results}
\label{sec_experimental_results}

In this section, we evaluate EAMDrift, encompassing various facets of its performance and capabilities. Firstly, we present a comparative analysis of our proposed method, and next, we present a study of its results.

\subsection{EAMDrift evaluation}

Table \ref{tab1} shows how our proposed model performed compared to other baseline and state-of-the-art methods on five different datasets. Each dataset was tested in two different steps. For example, for the NB1 dataset, we tested steps 1 and 7, which correspond to forecasting 1 and 7 days, respectively. Three columns are presented for the EAMDrift model, two for the non-interpretable versions that employ SVM and RDF as an ensemble (details in Section~\ref{experimenta_setup_details}), and the other for the original version of our proposed model, denoted as ``Original''. We did a previous grid search for the SARIMA and LSTM models to find the best parameters for each dataset.

\renewcommand{\arraystretch}{1.3}
\begin{table}
\centering
\caption{Comparison of MAPE values for different datasets using different models. A step column is presented for each dataset, meaning the number of points predicted. For the EAMDrift, three sub-columns are presented regarding the non-interpretable (``SVM" and ``RDF") and the original interpretable version (``Original").}\label{tab1}
\begin{tabular}{>{\arraybackslash}p{0.07\linewidth}|>{\centering\arraybackslash}p{0.08\linewidth}|>{\centering\arraybackslash}p{0.16\linewidth}>{\centering\arraybackslash}p{0.16\linewidth}|>{\centering\arraybackslash}p{0.16\linewidth}>{\centering\arraybackslash}p{0.16\linewidth}>{\centering\arraybackslash}p{0.16\linewidth}}
\arrayrulecolor{black}\specialrule{.11em}{0em}{0em} 
\multicolumn{2}{c|}{\textbf{Datasets}} &  \textbf{SARIMA} & \textbf{LSTM} & \multicolumn{3}{c}{\textbf{EAMDrift}}\\
\multicolumn{1}{c}{} & \multicolumn{1}{c|}{Step} &  & & \textbf{SVM} & \textbf{RDF} & \textbf{Original}\\
\arrayrulecolor{black}\specialrule{.11em}{0em}{0em} 

\cellcolor[gray]{0.95} & \cellcolor[gray]{0.95}1 & \cellcolor[gray]{0.95}30.45\% &\cellcolor[gray]{0.95} 33.82\% & \cellcolor[gray]{0.95}\textbf{16.92\%} & \cellcolor[gray]{0.95}18.46\% & \cellcolor[gray]{0.95}19.87\%  \\
\arrayrulecolor{lightgray}
\cline{2-7}
 \multirow{-2}{*}{\cellcolor[gray]{0.95} \rotatebox{90}{\textbf{NB1}}} & \cellcolor[gray]{0.95}7 & \cellcolor[gray]{0.95}31.68\% & \cellcolor[gray]{0.95}35.34\% &\cellcolor[gray]{0.95} \textbf{18.28\%} &\cellcolor[gray]{0.95} 21.12\% & \cellcolor[gray]{0.95}20.01\% \\
 \arrayrulecolor{black}\specialrule{.11em}{0em}{0em}

\cellcolor[gray]{1} & 1 & 49.11\% & 40.99\% & 26.87\% & \textbf{25.03\%} & 26.77\% \\
\arrayrulecolor{lightgray}
\cline{2-7}
 \multirow{-2}{*}{\cellcolor[gray]{1} \rotatebox{90}{\textbf{NB2}}} & 7 & 56.57\% & 39.46\% & \textbf{21.01\%} & 21.89\% & 23.78\% \\
\arrayrulecolor{black}\specialrule{.11em}{0em}{0em} 
\cellcolor[gray]{0.95}& \cellcolor[gray]{0.95}6 & \cellcolor[gray]{0.95}38.56\% &\cellcolor[gray]{0.95} 57.77\% & \cellcolor[gray]{0.95}28.19\% & \cellcolor[gray]{0.95}\textbf{27.32}\% &\cellcolor[gray]{0.95} 27.8\%  \\
\arrayrulecolor{lightgray}
\cline{2-7}
 \multirow{-2}{*}{\cellcolor[gray]{0.95} \rotatebox{90}{\textbf{GCT}}} & \cellcolor[gray]{0.95}12 &\cellcolor[gray]{0.95} 47.99\% &\cellcolor[gray]{0.95} 61.88\% & \cellcolor[gray]{0.95}\textbf{29.89\%} &\cellcolor[gray]{0.95} 33.97\% &\cellcolor[gray]{0.95} 31.09\% \\
\arrayrulecolor{black}\specialrule{.11em}{0em}{0em} 

 \cellcolor[gray]{1}& 12 &  60.91\% & 42.62\% & \textbf{15.32\%} & 16.31\% & 17.89\%  \\
\arrayrulecolor{lightgray}
\cline{2-7}
 \multirow{-2}{*}{\cellcolor[gray]{1} \rotatebox{90}{\textbf{EPC}}}  & 24 & 48.89\% & 46.58\% & \textbf{16.18\%} & 18.36\% & 19.54\% \\
\arrayrulecolor{black}\specialrule{.11em}{0em}{0em} 

\cellcolor[gray]{0.95} &\cellcolor[gray]{0.95} 1 &\cellcolor[gray]{0.95} 24.99\% &\cellcolor[gray]{0.95} 19.12\% & \cellcolor[gray]{0.95} 17.02\% & \cellcolor[gray]{0.95}16.67\% & \cellcolor[gray]{0.95}\textbf{16.41\%}  \\
\arrayrulecolor{lightgray}
\cline{2-7}
   \multirow{-2}{*}{\cellcolor[gray]{0.95} \rotatebox{90}{\textbf{MSV}}}  & \cellcolor[gray]{0.95}7 & \cellcolor[gray]{0.95}34.06\% & \cellcolor[gray]{0.95}30.87\% & \cellcolor[gray]{0.95}\textbf{22.15\%} & \cellcolor[gray]{0.95}23.11\% & \cellcolor[gray]{0.95}24.56\% \\
\arrayrulecolor{black}\specialrule{.11em}{0em}{0em}

\addlinespace[0.05cm]
\multicolumn{2}{l|}{\textbf{Mean Error}} & 42.32\% & 40.85\% & \textbf{21.18\%} & 22.22\% & 22.77\%  \\
\addlinespace[0.05cm]
\arrayrulecolor{black}\specialrule{.11em}{0em}{0em} 

\end{tabular}
\end{table}

Analyzing Table~\ref{tab1}, we can see that the baseline models had significantly higher errors than all versions of EAMDrift. This is because, although the models made predictions by training with data up to the prediction point, the parameters were only tuned once at the beginning, which likely worsened the results over time. In contrast, EAMDrift adjusts the parameters for each concept over time, allowing the model to automatically retrain and improve results.

Among all versions of EAMDrift, our proposed model had lower errors in just one test than the non-interpretable versions. The SVM version had the best mean error among all tests, but the difference to the other versions was only around 1-2\%. Therefore, while the non-interpretable versions of EAMDrift slightly improved outcomes, the differences in MAPE error were not significant enough to assume they were better. This highlights the effectiveness of EAMDrift and motivates the use of interpretability in time series analysis.

\subsection{Analysis of results}

\begin{figure}
\centering
\includegraphics[width=1\textwidth]{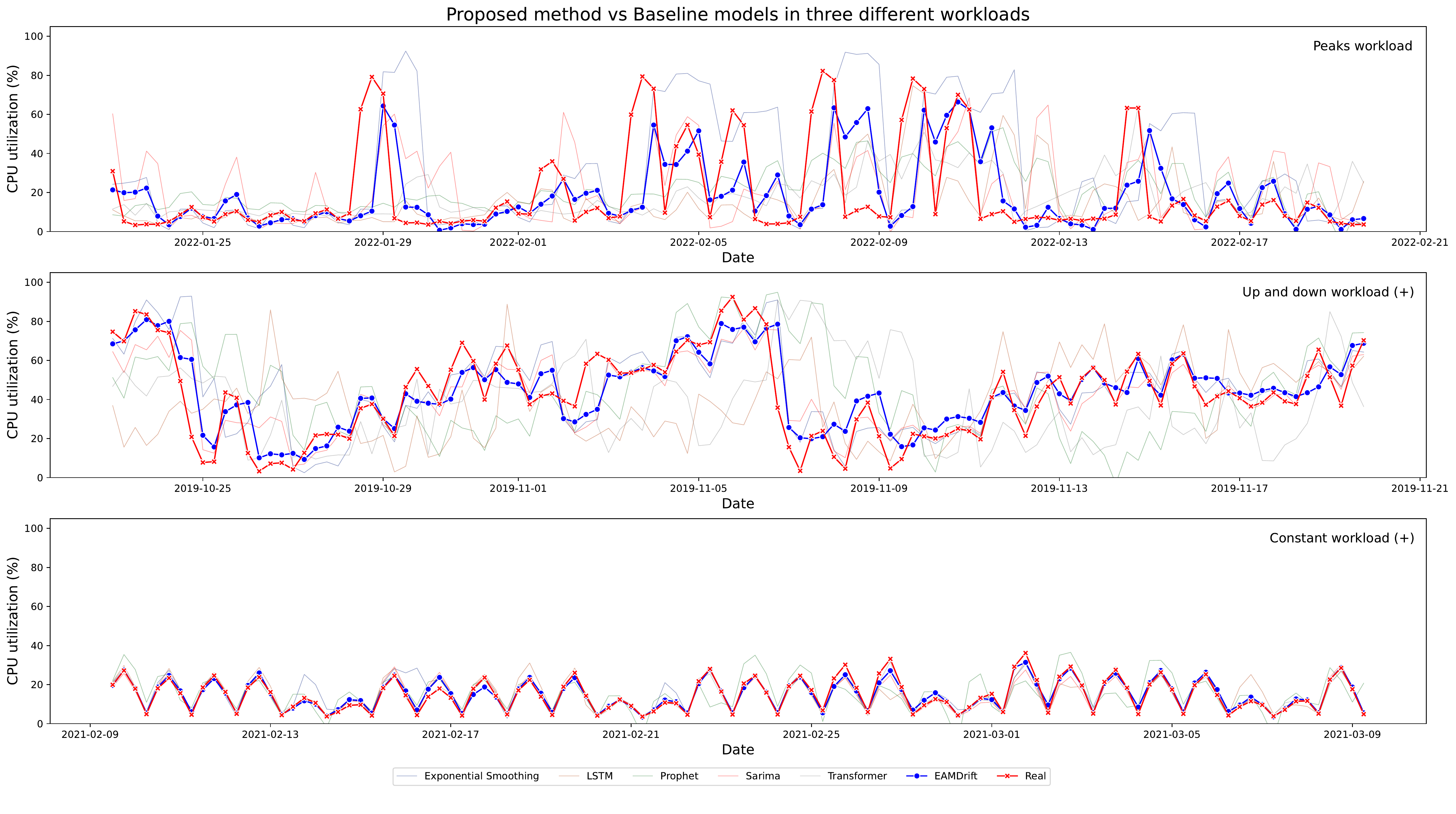}
\caption{Proposed method vs. individual predictors in three different workloads of NB1.} \label{fig:testingindifferentpatterns}
\end{figure}

To gain a better understanding of the results of EAMDrift, we selected three parts of the NB1 dataset. We analyzed the predictions made by our proposed model against the individual predictors used by our model. The results are depicted in Figure \ref{fig:testingindifferentpatterns}. Each graph represents a different concept (identified in the top right corner) and contains two thick lines --- one in red with the actual values and one in blue with the predictions made by our model. The remaining lines represent the baseline model's predictions. Our model provides better results when compared to the baseline model in all three patterns. Even in the first two patterns, where we observe a high presence of burst, our model adapted well. Furthermore, in the third plot, where a cyclic pattern is presented, all the models provided a good forecast, except for the Prophet model, which had some parts that yielded unreasonable results. Based on these three patterns, our model adapts well to bursts without sacrificing accuracy in simple patterns.

\begin{figure}
\centering
\includegraphics[width=1\textwidth]{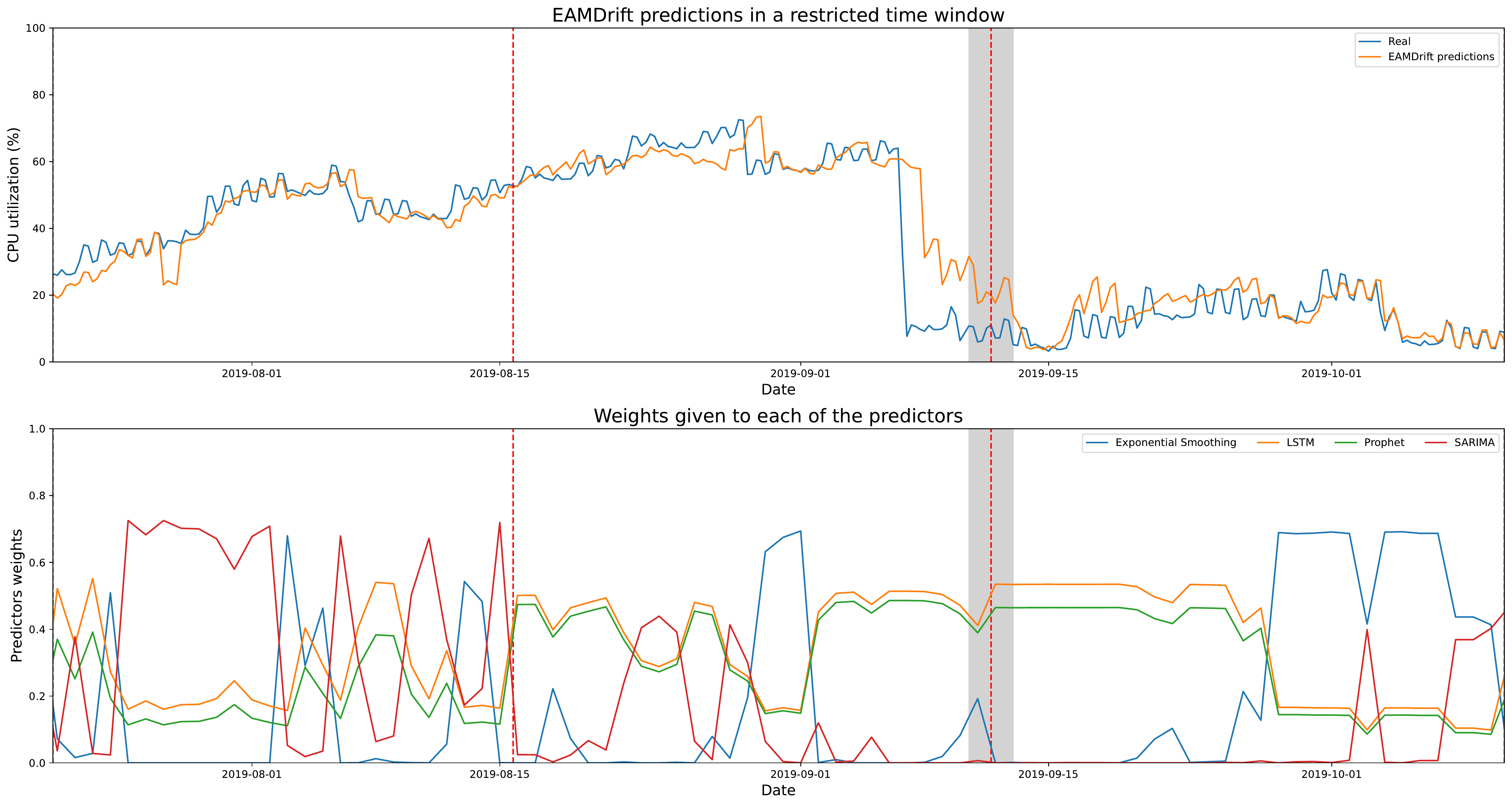}
\caption{Detailed analysis of our proposed model in a part of the NB1 dataset.} \label{fig:first_layer_simulation_highlighted}
\end{figure}

Next, in Figure~\ref{fig:first_layer_simulation_highlighted}, we highlight a part of the workload from the NB2 dataset to understand the predictions of EAMDrift better. The first plot presents the real and predicted workload, and the second shows the assigned weights to each model to obtain the final prediction. The Transformer model does not appear due to it being  discarded during training because it was not the best for at least more than one point. We note that the SARIMA model plays a significant role in the final predictions while the workload grows. When the workload decreases, SARIMA loses importance, and the remaining models take a different path.

In addition, the vertical red dashed lines in this Figure represent the zones detected by our drift to our model to be retrained. In the range of days presented in this Figure, the model was retrained twice, and each retraining period generated a new set of interpretable rules. For example, for the highlighted retrain point, one rule created was: \texttt{variance <= 7.2 AND sentiment\_overall <= 1.0 AND number\_negative\_tweets > 14.0 AND mean > 40}, which suggests a negative sentiment in zones where the \texttt{mean} is greater than 40\%.


\section{Conclusions}
\label{sec_conclusion}

Using machine learning prediction models in time series has shown tremendous potential in various research fields. In this work, we propose an innovative novel model called EAMDrift, which combines multiple models' strengths to produce more accurate and robust predictions. EAMDrift also has interpretable mechanisms that allow a better understanding of the predictions. This is particularly important for applications where precise forecasting is critical, such as financial markets and health problems.

To evaluate EAMDrift, we conducted comprehensive evaluations on different datasets over different models. The results show that our model outperforms the baseline ones by 20\%. Our model achieved comparable error rates compared to the best non-interpretable ensemble models. This suggests that interpretable machine learning models can be a viable solution for time series prediction.

Our results over different time series datasets were promising. We believe that the findings presented in this work can contribute to advancing the field of machine learning and inspire further research on bringing interpretability to time series forecasting. Finally, our methodology is easy to implement. A more detailed view of EAMDrift, its methods, and its usage (through a tutorial using the MSV dataset) can be seen in \url{https://anonymous.4open.science/r/EAMDrift-6DC0/README.md}.

\section{Ethical issues}



In this work, most data is publicly available and does not raise privacy concerns. However, we used news and user tweets to analyze sentiment and included them as covariates for some models. We want to clarify that we only used the tweet text and publication date for sentiment analysis. We discarded all text and ignored private information, such as user names or publication locations.

It is crucial to be cautious when working with data because no model is 100\% accurate. The responsibility for actions based on the model's predictions should be carefully considered. The data centers used in this study are an excellent example of this double-edged sword. While the forecasts can help allocate data center resources more effectively, saving energy and computing resources and offering better services by avoiding under-provisioning resources, predictions may also help hackers determine the best times to launch different attacks.

Furthermore, following the European ethics guidelines for trustworthy AI, our model provides interpretability for its predictions. This allows for transparency, confidence, and understanding of the predictions.

\end{document}